\pdfoutput=1

\documentclass[10pt,twocolumn,letterpaper]{article}

\usepackage{iccv}
\usepackage{times}
\usepackage{epsfig}
\usepackage{graphicx}
\usepackage{amsmath}
\usepackage{amssymb}
\usepackage{multirow}
\usepackage{appendix}

\usepackage[pagebackref=true,breaklinks=true,letterpaper=true,colorlinks,bookmarks=false]{hyperref}

\iccvfinalcopy 

\def\iccvPaperID{2432} 

\ificcvfinal\fi

\newcommand{\ParaSpace}{\vspace{-0.2mm}}

\begin{document}

\title{Enabling the Network to Surf the Internet\vspace{-0.2cm}}

\author{Zhuoling Li$^{1}$, Haohan Wang$^{1}$, Tymoteusz \'{S}wistek$^{1}$, Weixin Chen$^{1}$, Yuanzheng Li$^{2}$ Haoqian Wang$^{1}$\thanks{Corresponding author.}  \\
$^{1}$Tsinghua University $^{2}$Huazhong University of Science and Technology \\
{\tt\small \{lzl20, wang-hh20, tswistek20, chenwx20\}@mails.tsinghua.edu.cn yuanzheng\_li@hust.edu.cn} \\
{\tt\small haoqianwang@tsinghua.edu.cn}
}

\maketitle
\ificcvfinal\thispagestyle{empty}\fi

\begin{abstract}

Few-shot learning is challenging due to the limited data and labels. Existing algorithms usually resolve this problem by pre-training the model with a considerable amount of annotated data which shares knowledge with the target domain. Nevertheless, large quantities of homogenous data samples are not always available. To tackle this issue, we develop a framework that enables the model to surf the Internet, which implies that the model can collect and annotate data without manual effort. Since the online data is virtually limitless and continues to be generated, the model can thus be empowered to constantly obtain up-to-date knowledge from the Internet. Additionally, we observe that the generalization ability of the learned representation is crucial for self-supervised learning. To present its importance, a naive yet efficient normalization strategy is proposed. Consequentially, this strategy boosts the accuracy of the model significantly (20.46\% at most). We demonstrate the superiority of the proposed framework with experiments on miniImageNet, tieredImageNet and Omniglot. The results indicate that our method has surpassed previous unsupervised counterparts by a large margin (more than 10\%) and obtained performance comparable with the supervised ones.

\end{abstract}

\section{Introduction}

Although deep learning based methods have achieved prominent progress \cite{brown2020language, li2020deep, silver2016mastering, simonyan2014very}, their performance heavily relies on large-scale datasets with annotations \cite{hong2017weakly}. The significant cost of preparing numerous data arises the research interest towards few-shot learning \cite{snell2017prototypical}, which aims to train a model with only a few examples in each category. To realize this objective, the model is often first pre-trained using massive labeled data that shares knowledge with the target task. Then, through being fine-tuned on a handful of provided examples, the model could adapt to the target task quickly.
\begin{figure}[tb]
    \centering
    \includegraphics[scale=0.46]{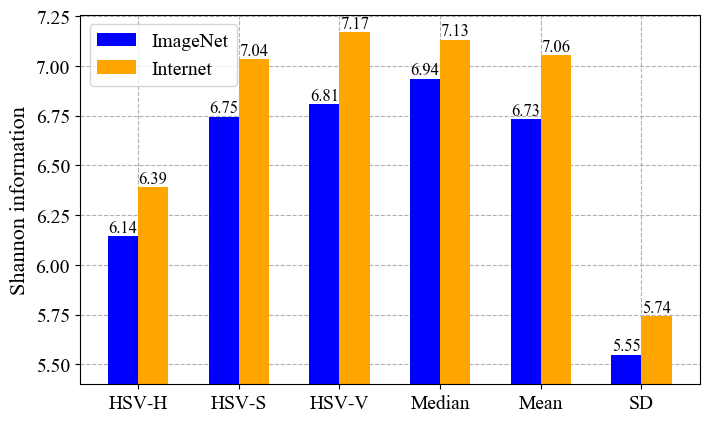}
    \caption{Shannon information of images between ImageNet and the web-crawled dataset under 6 statistical metrics. The y-axis reflects the Shannon information volume and the x-axis illustrates the 6 metrics. HSV-H, HSV-S and HSV-V are the mean values of pixels corresponding to three HSV color space channels. Median, Mean and SD denote the median value, mean value and standard deviation of the RGB color space pixels, respectively. Refer to Appendix A.1 for details.}
    \vspace{-0.4cm}
\end{figure}

Nevertheless, large volumes of annotated data related to the target task is rarely available. Many corner cases might not be included. Besides, although previous works argue that few-shot learning models benefit from more similar data \cite{chen2020big}, expanding existing homogeneous datasets takes great effort. These obstacles hinder the application of few-shot learning.

To address this issue, we raise a new question:

\vspace{1mm}
$\bullet$\textbf{$\ $ Is training a model without collecting and annotating data manually feasible?}
\vspace{1mm}

Recent progress in self-supervised learning brings hope to answering half of this question \cite{chen2020simple, he2020momentum}. By constructing pseudo labels, the process of annotating data manually could be avoided, but the trained model still produces representative embeddings.

In order to resolve the other half of the question, i.e., how to avoid collecting data manually, we seek help from the web images. Fortunately, a keyword-based image search can substitute the need to gather data manually. Furthermore, through devising self-supervised learning tasks on the web data, both the gathering and annotating data process could be excluded.

Moreover, we observe extra benefits of utilizing web-crawled data. As illustrated in Figure 1, we compare the Shannon information \cite{brukner2001conceptual} of images in ImageNet and the web-crawled dataset with respect to 6 statistical metrics. We find that web-crawled images contain more information than the ones in ImageNet. Hence, the web data seems to be more diverse than the manually prepared datasets. As data diversity favors self-supervised learning \cite{chen2020big}, utilizing the web data might be a promising alternative.

Incorporating the above insights, we develop a few-shot classification framework which first pre-trains the model through constructing pseudo labels on the web-crawled data, and then fine-tunes it based on a handful of support examples. To the best of our knowledge, this is the first technique that liberates people from manually preparing data for the few-shot classification pre-training phase.

Additionally, since the online data is virtually endless, the trained model could continually obtain knowledge from massive unseen images. In contrast to preceding lifelong learning methods restricted in a pre-defined scene \cite{parisi2019continual}, our framework addresses the wild data from the Internet.

Furthermore, we note from the experimental results that the generalization ability is crucial for self-supervised learning. To show its significance, a batch-instance normalization (BIN) strategy is devised and embedded into the networks. This strategy brings a significant accuracy gain (20.46\% at most, shown in Section 4.5).

In order to demonstrate the effectiveness of the proposed framework, we verify its performance on three benchmark datasets, miniImageNet \cite{vinyals2016matching}, tieredImageNet \cite{ren2018meta} and Omniglot \cite{lake2011one}. The results indicate that the proposed framework has surpassed the unsupervised counterparts significantly and obtained performance comparable to the supervised ones.

Comprehensively, our main contributions are summarized as follows:

\vspace{1mm}
$\bullet$$\ $We propose a few-shot classification framework named Surf on the Internet (SOI)\footnote{Code will be available online once the paper is accepted.}, which pre-trains models by constructing pseudo labels on web-crawled data without annotations. To the best of our knowledge, this is the first time that the manual operations are fully excluded from the preparing data procedure.

\vspace{1mm}
$\bullet$$\ $We highlight the importance of generalization ability for self-supervised learning. To present its significance, we devise a simple normalization strategy and embed it into the encoder networks. This strategy enhances the accuracy of the trained model with a large margin (20.46\% at most).

\vspace{1mm}
$\bullet$$\ $Evaluated on three benchmark datasets, the proposed framework has outperformed its unsupervised counterparts significantly (more than 10\%) and achieved accuracy comparable with the supervised methods.

\section{Related works}

\noindent \textbf{Few-shot classification.} Few-shot classification aims to distinguish the samples of the target tasks correctly based on a limited amount of labeled examples. According to the form of loss function, related methods could be divided into two categories, i.e., the metric-based and proxy-based methods \cite{sun2020circle}. Early works often employ the former one, the objective of which is to reinforce the similarity between the samples belonging to the same class \cite{allen2019infinite, hao2019collect, snell2017prototypical, sung2018learning, vinyals2016matching}. On the contrary, recently developed proxy-based methods aim to encode the input data into pre-set vectors \cite{dhillon2019baseline, rajasegaran2020self, tian2020rethinking}.

Following the proxy-based strategies, SOI optimizes the parameters of models with the target of minimizing the cross-entropy loss. Unlike preceding models only effective in restricted domains, SOI can handle manifold wild data.

\vspace{1mm}
\noindent \textbf{Self-supervised learning.} Without manual annotations, self-supervised methods train models through formulating tasks which involve the desired representation \cite{henaff2020data}. Commonly employed self-supervised tasks include colorizing gray-scale images \cite{larsson2017colorization}, frame order verification \cite{misra2016shuffle}, super resolution \cite{ledig2017photo} and some others \cite{chen2020generative, jing2020self}. Among the tasks designed for image classification, contrastive learning grabs great attention \cite{grill2020bootstrap, henaff2020data, khosla2020supervised}. It has been verified that the model pre-trained in the form of contrastive learning can produce very competitive representation \cite{li2020fewer}.

Nevertheless, contrastive learning is hardly studied under the few-shot classification setting. By contrast, SOI adopts it. Meanwhile, instead of utilizing the manually prepared datasets, SOI pre-trains models with only web-crawled images.

\vspace{1mm}
\noindent \textbf{Lifelong learning.} Studies about lifelong learning focus on how to obtain knowledge ceaselessly from a continuous sequence of tasks while avoiding the catastrophic forgetting problem \cite{lee2016dual, parisi2019continual, tessler2017deep}. Existing methods primarily conduct it in three patterns, which include replaying the old samples \cite{shin2017continual}, protecting the obtained information from being overwritten \cite{benna2016computational} and allocating more resources to store the new knowledge \cite{yoon2017lifelong}. Notably, almost all of them follow the same principle, i.e., while encountering new cases, all the historical knowledge should be retained to the most extent.

Contrary to aforementioned methods, SOI is driven by a conviction that not all of the previously encountered information is necessary and those cases from important domains appear more frequently. Therefore, SOI is philosophically different from preceding methods.
\begin{figure*}[htbp]
    \centering
    \includegraphics[scale=0.516]{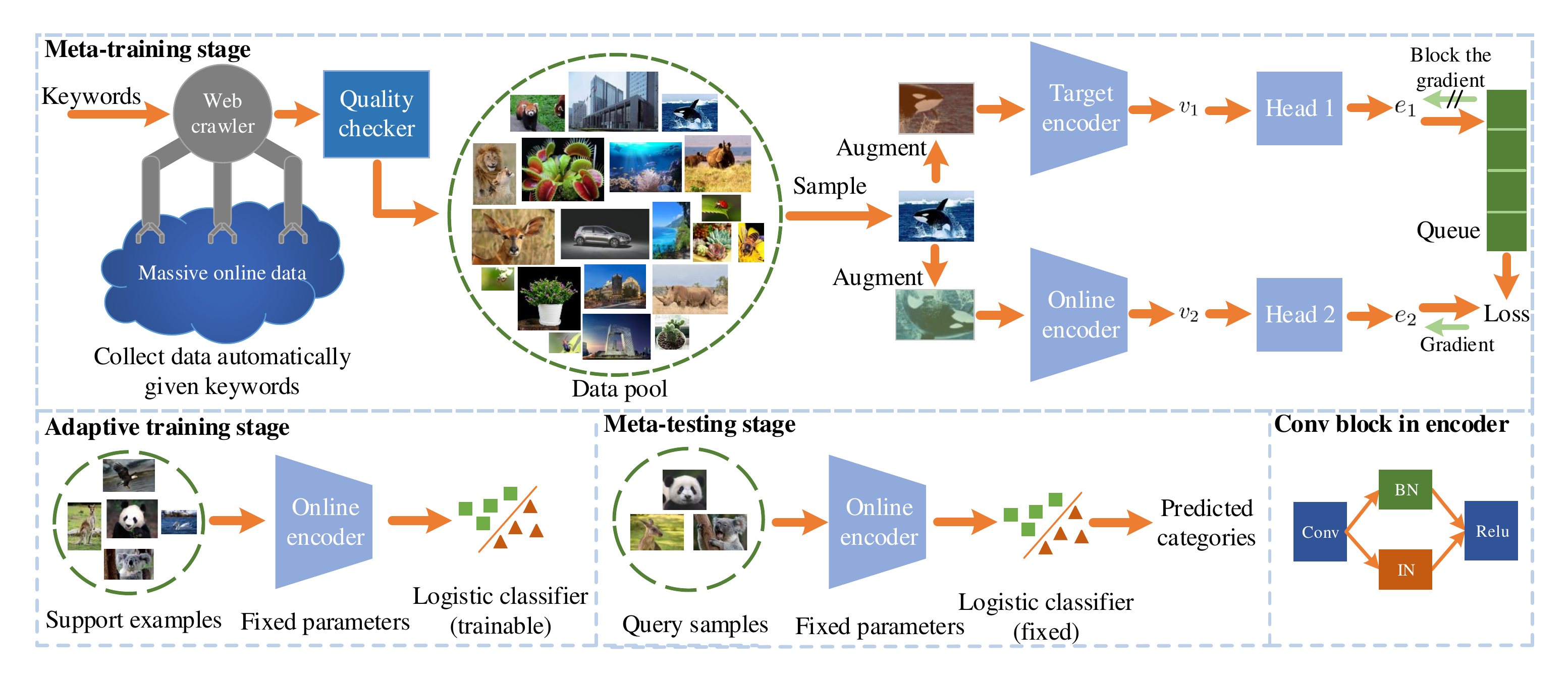}
    \caption{The overall framework of SOI (Conv: Convolution).}
    \vspace{-0.4cm}
\end{figure*}

\section{Framework}

In this section, we introduce how the proposed framework enables the network to surf the Internet. As shown in Figure 2, the framework includes three stages, i.e., the meta-training stage, adaptive training stage and meta-testing stage. In the first stage, large volume of data is downloaded from the Internet given the search keywords. Then, the encoders with BIN are trained in the form of contrastive learning to learn prior knowledge from the collected data. Afterwards, during the adaptive training stage, a logistic classifier is fine-tuned based on the support examples. Finally, in the meta-testing stage, combing the encoder and the logistic classifier, our method can classify the query samples with a competitive accuracy.

In the following, we present the details of the framework. First of all, Section 3.1 explains the overall pipeline. Then, Section 3.2$\sim$3.4 describe the highlights of the framework from three perspectives, respectively.

\ParaSpace
\subsection{Overall framework}
\ParaSpace

During the meta-training stage, a set of search keywords (e.g. plant, animal and building) are provided as the input to SOI. With these keywords, numerous images are downloaded by the web crawler. Then, a quality checker verifies the format of the images and removes the broken ones. The remaining images are shuffled and constitute a huge data pool.

We utilize the contrastive learning strategy \cite{he2020momentum} to construct pseudo labels and train the encoders. This strategy builds upon a simple assumption: the representation of an image is view-invariant. Thus, the representations of two independently and randomly augmented views should be similar. For instance, given two images $x_{a}$ and $x_{b}$, we can augment them randomly (e.g. crop or flip) and obtain the processed counterparts $x_{a}^{'}$ and $x_{b}^{'}$. If we treat $x_{a}$ as the query sample, $x_{a}^{'}$ and $x_{b}^{'}$ are the corresponding positive and negative samples, respectively. According to the above assumption, the representation of $x_{a}$ should be alike to $x_{a}^{'}$ while differing from $x_{b}^{'}$.

Therefore, for training a network $\Phi(\cdot)$, the optimization objective is to draw $\Phi(x_{a}^{'})$ closer to $\Phi(x_{a})$ and push away $\Phi(x_{b}^{'})$ from $\Phi(x_{a})$. In contrast to previous works which implement $\Phi(\cdot)$ with the Siamese network \cite{chen2020simple}, we employ two separate encoders with the same structure to extract the representation. In addition, we apply the BIN to encoders to enhance the generalization ability.

As depicted in Figure 2, after the augmentation operation, the two views are mapped to $v_{1}$ and $v_{2}$ by the two encoders (the target and online encoders), respectively. Then, $v_{1}$ and $v_{2}$ are further transformed into $e_{1}$ and $e_{2}$ by the two projection heads (head 1 and head 2). Since $e_{1}$ would be reused in subsequent training iterations, we push it into a dynamic queue. The optimization loss is computed based on $e_{2}$ and all vectors stored in the queue. The implementation details of the meta-training phase will be introduced in Section 3.2.

In the subsequent stages, parameters of the trained online encoder are frozen to extract the representation vectors. In order to group the vectors according to their categories correctly, a logistic classifier \cite{tian2020rethinking} is trained in the adaptive training stage. Afterwards, integrated with the frozen online encoder and the logistic classifier, our method can predict categories of the query samples during the meta-testing stage.

\ParaSpace
\subsection{Inductive contrastive learning}
\ParaSpace

Contrastive learning is a widely studied technique for constructing pseudo labels in recent unsupervised learning publications \cite{chen2020simple, chen2020big, grill2020bootstrap, he2020momentum}. However, few works considered using it under the few-shot setting. The most similar one to ours is few-BiT \cite{li2020fewer}. It employs contrastive learning technique under the transductive classification setting, meaning that the query sample distribution is observable while pre-training the model. Nevertheless, in many cases, query samples are not available in advance. Moreover, as a consequence of utilizing query samples during the pre-training stage, the obtained model is rendered task specific, thus requiring us to train a specific model for each target task. This procedure could result in huge resource waste.

Conversely, we treat the pre-training phase of SOI as an inductive optimization problem. Unlike few-BiT, SOI does not require any query sample in the meta-training phase. Therefore, the model trained under SOI can be shared by various target tasks.

Now, we introduce the implementation details of contrastive learning in SOI. Given an image $x$ sampled in the data pool, we first augment it into two different views, which are denoted as $x_{m}$ and $x_{n}$. The applied augmentation transformations are the same as SimCLR \cite{chen2020simple}, which include random crop, color distortion, Gaussian blur, etc. Then, $x_{m}$ and $x_{n}$ are encoded by the target and online encoders separately, which are formulated as follows:
\begin{align}
v_{1} = \Phi_{t}(x_{m};\omega_{\Phi_{t}}) \\
v_{2} = \Phi_{o}(x_{n};\omega_{\Phi_{o}})
\end{align}
where $\Phi_{t}(\cdot)$ and $\Phi_{o}(\cdot)$ are the target and online encoders, $\omega_{\Phi_{t}}$ and $\omega_{\Phi_{o}}$ denote their corresponding parameters, $v_{1}$ and $v_{2}$ represent the encoded vectors.

Afterwards, $v_{1}$ and $v_{2}$ are further mapped to $e_{1}$ and $e_{2}$ by two projection heads. Mathematically, it is expressed as follows:
\begin{align}
e_{1} = \psi_{t}(v_{1};\omega_{\psi_{t}}) \\
e_{2} = \psi_{o}(v_{2};\omega_{\psi_{o}})
\end{align}
where $\psi_{t}(\cdot)$ and $\psi_{o}(\cdot)$ denote the two projection heads, $\omega_{\psi_{t}}$ and $\omega_{\psi_{o}}$ represent their parameters. Next, $e_{1}$ is pushed into a dynamic queue, where the oldest is dequeued once a new element is enqueued.

Regarding $e_{2}$ as the query $q$ and denoting the $i_{\rm th}$ element in the queue as $p_{i}$, the optimization objective is to maximize the similarity between $q$ and $p_{0}$ ($p_{0}$ is also $e_{1}$), and minimize the similarity between $q$ and the other elements of the queue. Hence, the loss function could be defined as:
\begin{align}
L = -log \frac{exp(cos(q, p_{0}) / \tau)}{\sum\limits_{i=1}^{N} exp(cos(q, p_{i}) / \tau)}
\end{align}
where $cos(q, p_{i})$ represents the cosine similarity between $q$ and $p_{i}$, while $L$ and $\tau$ are the obtained loss value and the temperature hyper-parameter, respectively.

With respect to the target of minimizing $L$, we use SGD \cite{bottou2012stochastic} to update $\omega_{\Phi_{o}}$ and $\omega_{\psi_{o}}$. On the contrary, The gradient back-propagating towards $\Phi_{t}$ and $\psi_{t}$ is blocked. $\omega_{\Phi_{t}}$ and $\omega_{\psi_{t}}$ evolve following the momentum update rule \cite{he2020momentum}, which is presented as follows:
\begin{align}
\omega_{\Phi_{t}} = \eta \omega_{\Phi_{t}} + (1 - \eta) \omega_{\Phi_{o}} \\
\omega_{\psi_{t}} = \eta \omega_{\psi_{t}} + (1 - \eta) \omega_{\psi_{o}}
\end{align}
where $\eta$ is the momentum coefficient. Once the meta-training process is finished, we freeze $\omega_{\Phi_{o}}$ and remain $\Phi_{o}$ for the target tasks. Meanwhile, the other parts ($\Phi_{t}$, $\psi_{t}$ and $\psi_{o}$) are discarded.

\vspace{2mm}
\noindent \textbf{Is single encoder possible?} Aside from encoding $x_{m}$ and $x_{n}$ with separate encoders, another alternative is using the Siamese network \cite{chopra2005learning}, where the two encoders share the same parameters. However, previous study has suggested that this strategy yields poor performance \cite{he2020momentum}. Inspired by the reinforcement learning theory \cite{hasselt2010double}, we make intuitive sense that adopting two encoders with separate parameters boosts the accuracy through decoupling $v_{1}$ and $v_{2}$.

Specifically, assuming we encode $x_{m}$ and $x_{n}$ with a single network $\Phi(\cdot)$ whose parameter is denoted as $\omega$, the encoding procedure could be formulated as $v_{1}=\Phi(x_{m}|\omega)$ and $v_{2}=\Phi(x_{n}|\omega)$. Hence, $v_{1}$ is coupled with $v_{2}$ by $\omega$. Conversely, adopting two encoders with separate parameters contributes to decoupling $v_{1}$ from $v_{2}$. This issue could benefit the optimization process.

\ParaSpace
\subsection{Utilizing web-crawled data}
\ParaSpace

Training models with web-crawled data is an effective strategy to reduce the significant cost of collecting data. Nevertheless, this strategy is limited by two restrictions. First of all, the data from the Internet is noisy \cite{6015557}. The obtained images could be unrelated to the search keywords. Although many data cleaning strategies have been introduced, their effectiveness appears to be limited \cite{xiao2015learning}. Secondly, most methods utilizing web-crawled data still require manual annotations, and the collected data is only used to assist the training process, which implies that the manually prepared datasets (e.g. ImageNet \cite{deng2009imagenet}) are still necessary \cite{hong2017weakly, lee2019frame, pizzati2020domain, shen2018bootstrapping}.

SOI offers solutions to both of the two restrictions. First of all, although SOI collects data given the search keywords, it does not regard the keywords as the corresponding category labels. Hence, SOI does not assume any association between the keywords and the content of the obtained images. Secondly, the training process of SOI fully relies on the unlabeled web-crawled data, and still brings forth models with competitive accuracy.

\vspace{2mm}
\noindent \textbf{Lifelong learning.} Integrated with the above two advantages, SOI excludes human intervention from preparing data and training model completely. It allows the model to explore online images ceaselessly in line with the lifelong learning principles \cite{parisi2019continual}. Additionally, since online images are virtually limitless, the model trained by SOI could evolve continually as the data volume is expanded.

Notably, preceding studies about lifelong learning primarily focus on how to retain all the priorly obtained knowledge while conducting new tasks \cite{parisi2019continual}. Conversely, we argue that \textbf{it is not absolutely crucial to retain all previously learned information, and cases from important domains appear more frequently}. SOI follows this notion and concentrates on the valuable data automatically by design, as the often encountered images update the trained model with a higher frequency. Such learning strategy is consistent with how human beings obtain new knowledge. For instance, people tend to memorize the common knowledge while forgetting the incidental events. SOI aims to understand the general realistic data distribution rather than increasing the knowledge in a restricted domain. This is fundamentally different from the preceding methods \cite{cermelli2020modeling}. With the provided search keywords, the model trained under SOI can take advantage of the Internet resources to the greatest extent.

\ParaSpace
\subsection{Generalization ability matters}
\ParaSpace

We observed an unexpected phenomenon when we compared the performance of the Prototypical network \cite{snell2017prototypical} (Proto) and the SOI with Resnet50 \cite{he2016deep} as the encoder (S-Res50). As presented in Table 1, on the miniImageNet dataset \cite{vinyals2016matching}, S-Res50 outperforms Proto by 1.0\% ($50.4\% - 49.4\%$) and 3.0\% ($71.2\% - 68.2\%$) under the 1-shot and 5-shot settings, respectively. On the contrary, Proto beats S-Res50 on Omniglot with large margins, i.e., 25.9\% ($98.8\% - 72.9\%$) and 15.2\% ($99.7\% - 84.5\%$).

\begin{table}[ht]
    \vspace{-0.2cm}
    \centering
    \resizebox{80mm}{9mm}{
    \begin{tabular}{c|c|cc|cc}
    \hline \hline
    \multirow{2}{*}{Model} & \multirow{2}{*}{Supervision} & \multicolumn{2}{c}{miniImageNet} & \multicolumn{2}{c}{Omniglot} \\
    \cline{3-6}
    & & 1-shot & 5-shot & 1-shot & 5-shot \\
    \cline{1-6}
    Proto & Yes & 49.4\% & 68.2\% & 98.8\% & 99.7\% \\
    S-Res50 & No & 50.4\% & 71.2\% & 72.9\% & 84.5 \%  \\
    \hline \hline
    \end{tabular}}
    \caption{Performance contrast between Proto and S-Res50 on the miniImageNet and Omniglot datasets (5-way classification).}
    \vspace{-0.3cm}
\end{table}

To explore the reason, we contrasted the images in the two datasets. It was found that the web-crawled data is similar to the natural images in miniImageNet, while quite different from the handwritten characters in Omniglot. The domain gap leads to the performance discrepancy.

We could resolve this problem in two ways. First, SOI can work in the style of lifelong learning \cite{parisi2019continual} and obtain abundant knowledge from the web data, which might bridge the domain gap. The disadvantage of this strategy is an increase of computation burden. Alternatively, we can directly improve the generalization ability of the learned representation. Hence, \textbf{the generalization ability matters for self-supervised learning}.

In order to validate this observation, we devise the BIN. As declared in preceding publications, batch normalization (BN) prefers retaining the underlying features, such as brightness and texture, while instance normalization (IN) maintains more abstract information like the semantic content \cite{pan2018two}. Meanwhile, it has been proved that the abstract information generalizes better \cite{li2019clu}. Thus, a direct strategy is merging BN and IN into a single module, which results in BIN. Mathematically, BIN is formulated as follows:
\begin{align}
\Psi_{\rm BI}(x) = \gamma \cdot \Psi_{\rm B}(x) + (1 - \gamma) \cdot \Psi_{\rm I}(x)
\end{align}
where $x$ is the input, $\gamma$ represents the balance factor,  $\Psi_{\rm B}(\cdot)$, $\Psi_{\rm I}(\cdot)$ and $\Psi_{\rm BI}(\cdot)$ denote the BN, IN and BIN operators, respectively.

Although simple, BIN is remarkably efficient. As presented in Section 4.5, it enhances the accuracies of the trained models significantly (20.46\% at most). The results further confirm the significance of generalization ability for self-supervised learning.

\section{Experiment}

In this Section, we demonstrate the superiority of SOI using 6 experiments. Experiment 1 (Section 4.2) and experiment 2 (Section 4.3) confirm the effectiveness of SOI by contrasting its performance with that of previous unsupervised and supervised methods. The results indicate that SOI has outperformed the preceding unsupervised methods significantly and achieved accuracy on par with the supervised counterparts. Then, experiment 3 (Section 4.4) and experiment 4 (Section 4.5) analyze the impact of the crawled data volume and BIN. The consequence proves the value of incremental data and claims that BIN can enhance the generalization ability of the learned representation significantly. Moreover, in experiment 5 (Section 4.6), the influence of various classifiers is studied. Finally, experiment 6 (Section 4.7) visualizes the generated embeddings from the model and reveals the discrepancy between unsupervised learning and supervised learning. Besides, the experimental details are introduced in Section 4.1.

\ParaSpace
\subsection{Settings}
\ParaSpace

\noindent \textbf{Data description.} The data utilized in the experiments is composed of two parts, the web-crawled data and three benchmark datasets, i.e., miniImageNet, tieredImageNet and Omniglot.

The web data is aggregated by using 267 search keywords, each of which represents a name of an object, e.g., koala, castle, doctor, etc. For each provided keyword, about 3000 corresponding images are downloaded from the Internet. The total storage space taken by the web-crawled data is 85G. The meta-training stage of SOI fully relies on these web-crawled images.

We compare the performance of the models on the benchmark datasets (miniImageNet, tieredImageNet and Omniglot) under the $n$-way $k$-shot protocol \cite{dhillon2019baseline}. For each iteration, the model needs to distinguish $n$ classes of query samples with the help of $k$ support examples per category. The split of the benchmark datasets follows the previous works \cite{dhillon2019baseline, snell2017prototypical}.

\vspace{2mm}
\noindent \textbf{Model description.} In the experiments, we analyze the effectiveness of SOI with various encoders, which include Resnet12, Resnet18 and Resnet50 \cite{he2016deep}. For simplicity, $\gamma$ is set as 0.5. The other hyper-parameters of the encoders, such as the number of convolution channels, stay consistent with the preceding methods \cite{dhillon2019baseline, tian2020rethinking}. 

\begin{table}[htbp]
    \vspace{-0.2cm}
    \centering
    \resizebox{80mm}{30mm}{
    \begin{tabular}{c|cccc}
    \hline \hline
    \multirow{2}{*}{Model} &  \multicolumn{4}{c}{Accuracy (way, shot)} \\
    \cline{2-5}
    & (5, 1) & (5, 5) & (5, 20) & (5, 50) \\
    \cline{1-5}
    BiGAN-KNN & 25.56\% & 31.10\% & 37.31\% & 43.60\% \\
    BiGAN-LC & 27.08\% & 33.91\% & 44.00 \% & 50.41\% \\
    BiGAN-MLP & 22.91\% & 29.06\% & 40.06\% & 48.36\% \\
    BiGAN-CLM & 24.63\% & 29.49\% & 33.89\% & 36.13\% \\
    BiGAN-CM & 36.24\% & 51.28\% & 61.33\% & 66.91\% \\
    BiGAN-CP & 36.62\% & 50.16\% & 59.56\% & 63.27\% \\
    \cline{1-5}
    DeepCluster-KNN & 28.90\% & 42.25\% & 56.44\% & 63.90\% \\
    DeepCluster-LC & 29.44\% & 39.79\% & 56.19\% & 65.28\% \\
    DeepCluster-MLP & 29.03\% & 39.67\% & 52.71\% & 60.95\% \\
    DeepCluster-CLM & 22.20\% & 23.50\% & 24.97\% & 26.87\% \\
    DeepCluster-CM & 39.90\% & 53.97\% & 63.84\% & 69.64\% \\
    DeepCluster-CP & 39.18\% & 53.36\% & 61.54\% & 63.55\% \\
    \cline{1-5}
    SOI-Res12 (ours) & 45.35\% & 63.06\% & 74.53\% & 78.87\% \\
    SOI-Res18 (ours) & 49.85\% & 68.59\% & 78.05\% & 82.04\% \\
    SOI-Res50 (ours) & 51.06\% & 72.62\% & 81.30\% & 85.77\% \\
    \hline \hline
    \end{tabular}}
    \caption{Comparison with preceding unsupervised methods on miniImageNet.}
    \vspace{-0.3cm}
\end{table}

\begin{table}[htbp]
    \vspace{-0.2cm}
    \centering
    \resizebox{80mm}{30mm}{
    \begin{tabular}{c|cccc}
    \hline \hline
    \multirow{2}{*}{Model} &  \multicolumn{4}{c}{Accuracy (way, shot)} \\
    \cline{2-5}
    & (5, 1) & (5, 5) & (20, 1) & (20, 5) \\
    \cline{1-5}
    BiGAN-KNN & 49.55\% & 68.06\% & 27.37\% & 46.70\% \\
    BiGAN-LC & 48.28\% & 68.72\% & 27.80\% & 45.82\% \\
    BiGAN-MLP & 40.54\% & 62.56\% & 19.92\% & 40.71\% \\
    BiGAN-CLM & 43.96\% & 58.62\% & 21.54\% & 31.06\% \\
    BiGAN-CM & 58.18\% & 78.66\% & 35.56\% & 58.62\% \\
    BiGAN-CP & 54.74\% & 71.69\% & 33.40\% & 50.62\% \\
    \cline{1-5}
    ACAI/DC-KNN & 57.46\% & 81.16\% & 39.73\% & 66.38\% \\
    ACAI/DC-LC & 61.08\% & 81.82\% & 43.20\% & 66.33\% \\
    ACAI/DC-MLP & 51.95\% & 77.20\% & 30.65\% & 58.62\% \\
    ACAI/DC-CLM & 54.94\% & 71.09\% & 32.19\% & 45.93\% \\
    ACAI/DC-CM & 68.84\% & 87.78\% & 48.09\% & 73.36\% \\
    ACAI/DC-CP & 68.12\% & 83.58\% & 47.75\% & 66.27\% \\
    \cline{1-5}
    SOI-Res12 (ours) & 80.00\% & 92.11\% & 57.76\% & 75.03\% \\
    SOI-Res18 (ours) & 80.05\% & 93.62\% & 58.99\% & 77.21\% \\
    SOI-Res50 (ours) & 74.08\% & 87.26\% & 48.64\% & 61.93\% \\
    \hline \hline
    \end{tabular}}
    \caption{Comparison with preceding unsupervised methods on Omniglot.}
    \vspace{-0.3cm}
\end{table}

\ParaSpace
\subsection{Comparison with unsupervised methods}
\ParaSpace

We compare our method with BiGAN \cite{donahue2019large}, DeepCluster \cite{tian2017deepcluster} and ACAI \cite{berthelot2018understanding}. The experiments are conducted on a natural image dataset (miniImageNet) and a handwritten character dataset (Omniglot). Table 2 and Table 3 report the results. The abbreviations in the tables, i.e., KNN, LC, MLP, CLM, CM and CP, represent K-nearest neighbors, linear classifier, multi-layer perception, cluster matching, CACTUs-MAML and CACTUs-ProtoNets \cite{hsu2018unsupervised}, respectively. SOI-Res12, SOI-Res18 and SOI-Res50 denote the SOI that adopts Resnet12, Resnet18 and Resnet50 \cite{he2016deep} as the encoder.

From the results in Table 2, we are able to draw two main conclusions. First of all, the SOIs with various encoders beat its unsupervised counterparts on miniImageNet with large margins. For instance, as seen in the $14_{\rm th}$ and $17_{\rm th}$ rows of Table 2, SOI-Res50 outperforms DeepCluster-CP by 11.88\% ($51.06\% - 39.18\%$), 19.26\% ($72.62\% - 53.36\%$), 19.76\% ($81.30\% - 61.54\%$) and 22.22\% ($85.77\% - 63.55\%$) regarding the 4 evaluation settings (5-way 1-shot, 5-way 5-shot, 5-way 20-shot and 5-way 50-shot). Secondly, as shown in Table 2, SOI-Res50 behaves significantly better than SOI-Res12 and SOI-Res18, which suggests that the accuracy would be enhanced even further with the growing number of the trained encoder parameters.

According to Table 3, in general, SOI surpasses the compared methods on Omniglot. Nevertheless, the obtained accuracies of SOI-Res12 and SOI-Res18 are higher than that of SOI-Res50. We suspect that the domain gap between the natural images and handwritten character pictures leads to this phenomenon. Specifically, the pictures contained in Omniglot differ considerably from the web-crawled data, causing the encoder with larger number of parameters to be affected more seriously by over-fitting. Hence, SOI-Res50 performs worse because it contains more parameters.

\begin{table}[htbp]
    \vspace{-0.2cm}
    \centering
    \resizebox{70mm}{35mm}{
    \begin{tabular}{c|cc}
    \hline \hline
    \multirow{2}{*}{Model} &  \multicolumn{2}{c}{Accuracy (\%)} \\
    \cline{2-3}
    & 5-way, 1-shot & 5-way, 5-shot \\
    \cline{1-3}
    MAML  & 48.70$\pm$1.84 & 63.11$\pm$0.92 \\
    Matching networks & 43.56$\pm$0.84 & 55.31$\pm$0.73 \\
    IMP & 49.2$\pm$0.7 & 64.7$\pm$0.7 \\
    Prototypical networks & 49.42$\pm$0.78 & 68.20$\pm$0.66 \\
    TAML & 51.77$\pm$1.86 & 66.05$\pm$0.85 \\
    SAML & 52.22$\pm$n/a & 66.49$\pm$n/a \\
    GCR & 53.21$\pm$0.80 & 72.34$\pm$0.64 \\
    KTN (Visual) & 54.61$\pm$0.80 & 71.21$\pm$0.66 \\
    PARN & 55.22$\pm$0.84 & 71.55$\pm$0.66 \\
    Dynamic few-shot & 56.20$\pm$0.86 & 73.00$\pm$0.64 \\
    Relation networks & 50.44$\pm$0.82 & 65.32$\pm$0.70 \\
    R2D2 & 51.2$\pm$0.6 & 68.8$\pm$0.1 \\
    SNAIL &  55.71$\pm$0.99 & 68.88$\pm$0.92 \\
    AdaResNet & 56.88$\pm$0.62 & 71.94$\pm$0.57 \\
    \cline{1-3}
    SOI-Res12 (ours) & 45.35$\pm$0.74 & 63.06$\pm$0.63 \\
    SOI-Res18 (ours) & 49.85$\pm$0.84 & 68.59$\pm$0.62 \\
    SOI-Res50 (ours) & 51.06$\pm$0.77 & 72.62$\pm$0.66 \\
    \hline \hline
    \end{tabular}}
    \caption{Comparison with preceding supervised methods on miniImageNet.}
    \vspace{-0.3cm}
\end{table}

\ParaSpace
\subsection{Comparison with supervised methods}
\ParaSpace

In this part, we compare SOI against some supervised few-shot classification methods on miniImageNet and tieredImageNet. The contrasted ones include MAML \cite{finn2017model}, Matching networks \cite{vinyals2016matching}, IMP \cite{allen2019infinite}, Prototypical networks \cite{snell2017prototypical}, TAML \cite{jamal2019task}, SAML \cite{hao2019collect}, GCR \cite{li2019few}, KTN (Visual) \cite{peng2019few}, PARN \cite{wu2019parn}, Dynamic few-shot \cite{gidaris2018dynamic}, Relation networks \cite{sung2018learning}, R2D2 \cite{bertinetto2018meta}, SNAIL \cite{mishra2017simple} and AdaResNet \cite{munkhdalai2018rapid}. The results on two datasets are summarized in Table 4 and Table 5, respectively. To be consistent with the results in previous works, we present both the mean classification accuracy and 95\% confidence interval.

\begin{table}[htbp]
    \vspace{-0.2cm}
    \centering
    \resizebox{70mm}{16mm}{
    \begin{tabular}{c|cc}
    \hline \hline
    \multirow{2}{*}{Model} &  \multicolumn{2}{c}{Accuracy (\%)} \\
    \cline{2-3}
    & 5-way, 1-shot & 5-way, 5-shot \\
    \cline{1-3}
    MAML  & 51.67$\pm$1.81 & 70.30$\pm$1.75 \\
    Prototypical networks  & 53.31$\pm$0.89 & 72.69$\pm$0.74 \\
    Relation networks  & 54.48$\pm$0.93 & 71.32$\pm$0.78 \\
    \cline{1-3}
    SOI-Res12 (ours) & 52.42$\pm$0.82 & 68.93$\pm$0.76 \\
    SOI-Res18 (ours) & 54.81$\pm$0.80 & 71.55$\pm$0.71 \\
    SOI-Res50 (ours) & 58.29$\pm$0.77 & 75.17$\pm$0.76 \\
    \hline \hline
    \end{tabular}}
    \caption{Comparison with preceding supervised methods on tieredImageNet.}
    \vspace{-0.3cm}
\end{table}

From Table 4 and Table 5, we note that the performance of SOI is on par with the supervised models. For example, the Prototypical network is a well-known baseline of few-shot classification. Validated on miniImageNet, its corresponding accuracies under 5-way 1-shot and 5-way 5-shot settings are 49.42\% and 68.20\%, respectively. By contrast, SOI-Res50 obtains the accuracies of 51.06\% and 72.62\%. The results indicate that SOI-Res50 has surpassed the prototypical networks. Moreover, although some supervised methods still outperform SOI slightly, such as AdaResNet, it can be regarded as a cost and performance trade-off.

\ParaSpace
\subsection{Analysis of crawled data volume}
\ParaSpace

This section aims to analyze how the crawled data volume influences the final classification accuracy. To that end, we utilize different percentages of the total crawled data to train the model. Resnet18 is employed as the encoder and we validate its performance on miniImageNet with respect to 5 evaluation settings (1-shot, 3-shot, 5-shot, 10-shot and 20-shot). The results are illustrated in Figure 3.

\begin{figure}[htbp]
    \vspace{-0.2cm}
    \centering
    \includegraphics[scale=0.5]{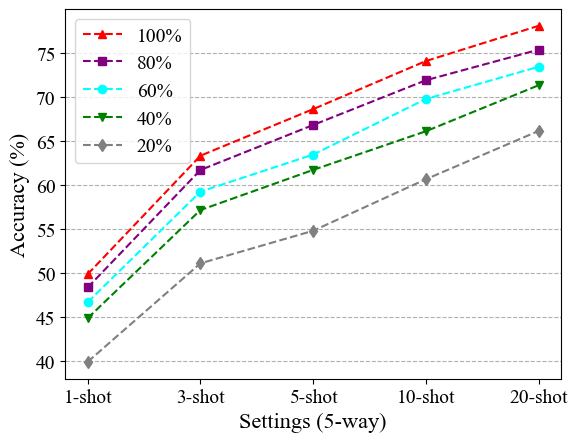}
    \caption{The influence of the crawled data volume.}
    \vspace{-0.2cm}
\end{figure}

We could draw two conclusions from Figure 3. First, the classification performance is improved with the increasement of the support examples. Therefore, we can enhance the accuracy through simply incorporating more support examples. Second, the crawled data volume benefits contrastive learning significantly. This observation further reveals the advantage of SOI, which can obtain fresh knowledge from the endless web data automatically.

\ParaSpace
\subsection{Ablation study of BIN}
\ParaSpace

In order to prove the value of generalization ability for self-supervised learning and verify the effectiveness of BIN, we compare the performance of the SOI with BIN and the SOI without BIN on miniImageNet and Omniglot. Three encoders (Resnet12, Resnet18 and Resnet50) are severally adopted to conduct the experiments. The results are reported in Table 6 and Table 7.

\begin{table}[htbp]
    \vspace{-0.2cm}
    \centering
    \resizebox{78mm}{16mm}{
    \begin{tabular}{c|c|cccc}
    \hline \hline
    \multirow{2}{*}{Use BIN} & \multirow{2}{*}{Encoder} &  \multicolumn{4}{c}{Accuracy (5-way)} \\
    \cline{3-6}
    & & 1-shot & 5-shot & 20-shot & 50-shot \\
    \cline{1-6}
    \multirow{3}{*}{No} & Res12 & 42.48\% & 58.38\% & 69.81\% & 75.08\% \\
    & Res18 & 46.01\% & 62.32\% & 72.60\% & 77.36\% \\
    & Res50 & 50.44\% & 71.20\% & 79.85\% & 84.13\% \\
    \cline{1-6}
    \multirow{3}{*}{Yes} & Res12 & 45.35\% & 63.06\% & 74.53\% & 78.87\% \\
    & Res18 & 49.85\% & 68.59\% & 78.05\% & 82.04\% \\
    & Res50 & 51.06\% & 72.62\% & 81.30\% & 85.77\% \\
    \hline \hline
    \end{tabular}}
    \caption{Ablation study of BIN on miniImageNet.}
    \vspace{-0.3cm}
\end{table}

As shown, BIN boosts the accuracy of SOI significantly. For instance, according to the $3_{\rm rd}$ and $6_{\rm th}$ rows of Table 7, under the 4 evaluation settings (5-way 1-shot, 5-way 5-shot, 20-way 1-shot and 20-way 5-shot), BIN enhances the accuracy of SOI-Res12 by 8.28\% ($80.00\% - 71.72\%$), 7.94\% ($92.11\% - 84.17\%$), 16.53\% ($57.76\% - 41.23\%$) and 20.46\% ($75.03\% - 54.57\%$) on Omniglot. Correspondingly, the performance improvements of SOI-Res12 on miniImageNet are 2.87\% ($45.35\% - 42.48\%$), 4.68\% ($63.06\% - 58.38\%$), 4.72\% ($74.53\% - 69.81\%$) and 3.79\% ($78.87\% - 75.08\%$).

\begin{table}[htbp]
    \vspace{-0.2cm}
    \centering
    \resizebox{78mm}{16mm}{
    \begin{tabular}{c|c|cccc}
    \hline \hline
    \multirow{2}{*}{Use BIN} & \multirow{2}{*}{Encoder} &  \multicolumn{4}{c}{Accuracy (way, shot)} \\
    \cline{3-6}
    & & (5, 1) & (5, 5) & (20, 1) & (20, 5) \\
    \cline{1-6}
    \multirow{3}{*}{No} & Res12 & 71.72\% & 84.17\% & 41.23\% & 54.57\% \\
    & Res18 & 75.06\% & 87.52\% & 46.11\% & 59.99\% \\
    & Res50 & 72.85\% & 84.49\% & 39.98\% & 54.05\% \\
    \cline{1-6}
    \multirow{3}{*}{Yes} & Res12 & 80.00\% & 92.11\% & 57.76\% & 75.03\% \\
    & Res18 & 80.05\% & 93.62\% & 58.99\% & 77.21\% \\
    & Res50 & 74.08\% & 87.26\% & 48.64\% & 61.93\% \\
    \hline \hline
    \end{tabular}}
    \caption{Ablation study of BIN on Omniglot.}
    \vspace{-0.3cm}
\end{table}

\begin{figure*}[htbp]
    \centering
    \includegraphics[scale=0.35]{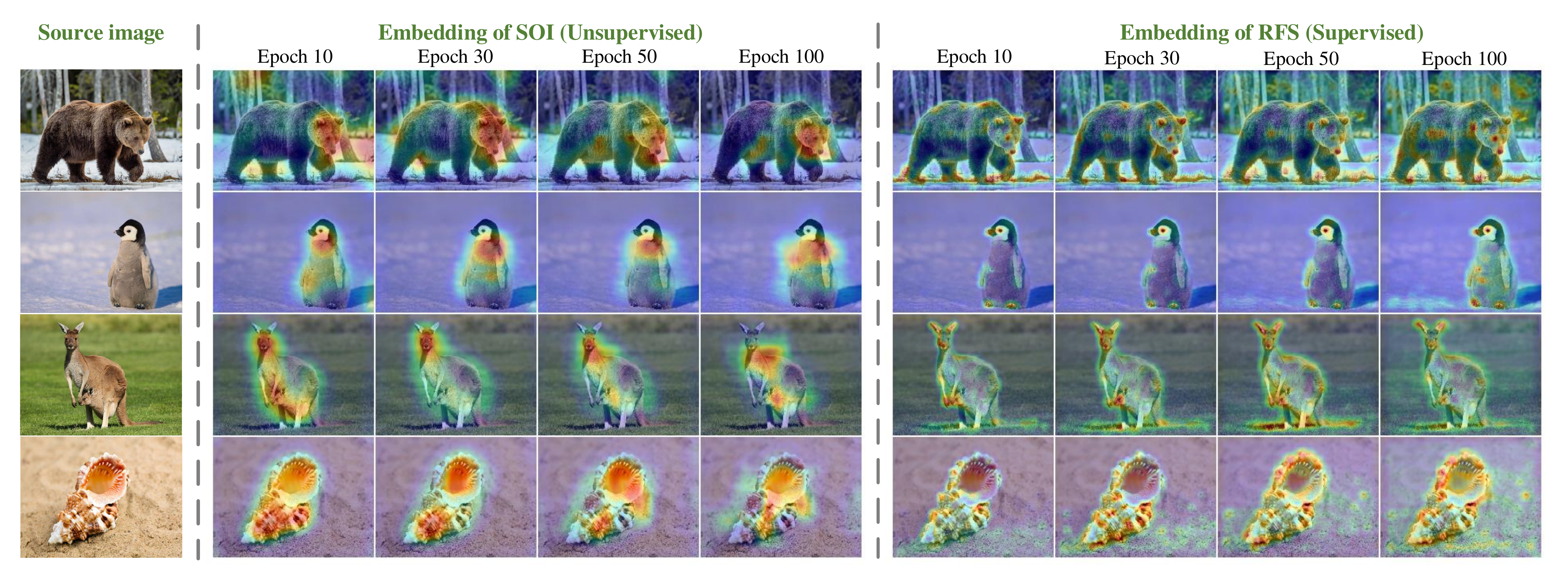}
    \caption{Embedding comparison between SOI (unsupervised) and RFS (supervised).}
    \vspace{-0.4cm}
\end{figure*}

Observing from Table 6 and Table 7, BIN brings greater improvements on Omniglot than miniImageNet. We assume this is because that the gap between the web-crawled dataset and Omniglot is larger in comparison with miniImageNet, and BIN successfully addresses it. This issue further affirms the effectiveness of BIN on bridging the domain gap and enhancing the generalization ability.

\ParaSpace
\subsection{Impact of adopting various classifiers}
\ParaSpace

As mentioned in Section 3.1, during the adaptive training stage, we fine-tune a classifier based on the support examples. In this experiment, we analyze how different classifiers influence the final classification accuracy on miniImageNet. The tested classifiers include LR (logistic regression), SVM (support vector machine), NN (nearest classifier), Cosine (cosine classifier) and Proto (prototype classifier) \cite{michie1994machine}. We present the accuracy and 95\% confidence interval in Table 8.

\begin{table}[ht]
    \vspace{-0.2cm}
    \centering
    \resizebox{70mm}{14mm}{
    \begin{tabular}{c|c|cc}
    \hline \hline
    \multirow{2}{*}{Model} & \multirow{2}{*}{Classifier} & \multicolumn{2}{c}{Accuracy (\%)} \\
    \cline{3-4}
    & & 5-way 1-shot & 5-way 5-shot \\
    \cline{1-4}
    \multirow{5}{*}{SOI-Res18} & LR & 49.85 $\pm$ 0.84 & 68.59 $\pm$ 0.62 \\
    & SVM & 49.02 $\pm$ 0.80 & 68.69 $\pm$ 0.73 \\
    & NN & 48.83 $\pm$ 0.75 & 62.12 $\pm$ 0.71 \\
    & Cosine & 49.44 $\pm$ 0.72 & 61.47 $\pm$ 0.73 \\
    & Proto & 48.34 $\pm$ 0.85 & 66.84 $\pm$ 0.78 \\
    \hline \hline
    \end{tabular}}
    \caption{The impact of adopting various classifiers on miniImageNet.}
    \vspace{-0.3cm}
\end{table}

As shown, by adopting Resnet18 as the online encoder, LR achieves the highest accuracy (49.85\%) under the 5-way 1-shot setting, and Proto behaves the worst (48.34\%). For the 5-way 5-shot setting, SVM presents the most accurate classification results (68.69\%), while the performance of Cosine is the poorest (61.47\%). Meanwhile, we could observe that the accuracy of Proto (68.59\%) is on par with SVM (68.69\%) under the 5-way 5-shot setting. Thus, LR performs the best.

\ParaSpace
\subsection{Embedding visualization}
\ParaSpace

In this section, we compare the embeddings generated by SOI and RFS \cite{tian2020rethinking} (a representative supervised SOTA of few-shot classification) to study the difference between unsupervised and supervised methods. Figure 4 presents the class activation maps output by the final convolution layer of SOI and RFS at different training epochs. The activated areas are where the network concentrates on. We find significant difference between the representations from SOI and RFS, which reveals the huge gap between unsupervised and supervised learning. As shown in Figure 4, SOI focuses more on the critical regions for classification, while RFS is distracted by the background areas. This phenomenon implies the superiority of SOI.

\begin{figure}[htbp]
    \centering
    \includegraphics[scale=0.44]{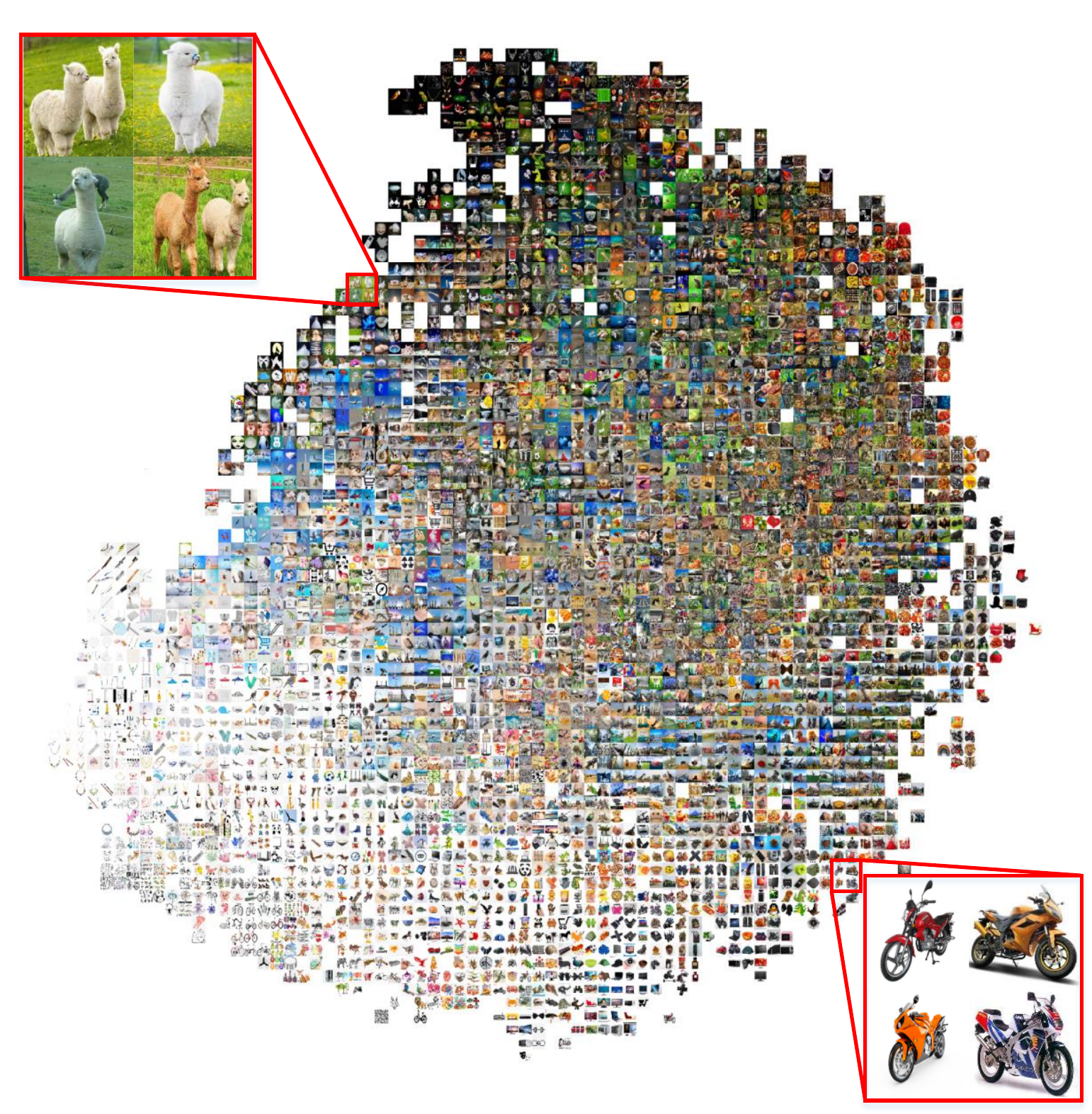}
    \caption{Visualization of the representation calculated by SOI with t-SNE on the web-crawled data.}
    \vspace{-0.3cm}
\end{figure}

Furthermore, we visualize the representation obtained from SOI using the t-SNE algorithm \cite{maaten2008visualizing} on the web-crawled dataset. As shown in Figure 5, the objects of the same or similar categories are well grouped. The results prove the promising clustering ability of SOI.

\section{Conclusion}

In this paper, we have proposed a framework which enables the model to surf the Internet. Without the requirement of collecting and annotating data manually, this framework first pre-trains a deep learning model through constructing pseudo labels on the web-crawled data. Then, it fine-tunes the model based on a few support examples. Based on this simple yet effective strategy, the proposed framework has outperformed preceding unsupervised counterparts by large margins and obtained results comparable with the supervised methods. Our experiment results have proved the value of this framework and indicated that further improvements would be obtained through increasing the amount of crawled data. We believe that this framework contributes to a solution of making the deep learning model understand the general knowledge about the real world.

\begin{appendix}

\renewcommand{\appendixname}{Appendix~\Alph{section}}

\section{Appendix}

\subsection{Details of Shannon information in Figure 1}

Here, we describe how to compute the Shannon information illustrated in Figure 1.

In order to measure the Shannon information volume in a dataset $\mathcal{D}$ with $N$ images, we first need to represent each image of $\mathcal{D}$ with a single number. We have employed 6 statistical metrics to generate this number, which include HSV-H, HSV-S, HSV-V, Median, Mean and SD. Specifically, HSV-H, HSV-S and HSV-V are the mean values of the pixels in the three HSV color space channels. Median, Mean and SD denote the median value, mean value and standard deviation of the RGB color space pixels, respectively. Adopting any one of these 6 metrics, we can obtain a numerical sequence with the length of $N$ to represent $\mathcal{D}$.

Then, we convert the elements in the sequence to integers. Since the numerical range of the image pixels is 0$\sim$ 255, there are 256 potential values for the converted elements of the sequence.

Afterwards, we count the occurring probability of every potential value and store it as an array with the length of 256. Denoting the $i_{\rm th}$ element of the array as $p_{i}$, the process of calculating the Shannon information volume could be formulated as:
\begin{equation}
H = -\sum\limits_{i=1}^{256} p_{i}log_{2}p_{i}
\end{equation}
where $H$ is the Shannon information amount of the dataset $\mathcal{D}$.

\end{appendix}

{\small
\bibliographystyle{ieee_fullname}
\bibliography{references}
}

\end{document}